%% file: main.tex
\pdfoutput=1
\def\year{2020}\relax
\documentclass[letterpaper]{article} 
\usepackage{aaai20}  
\usepackage{times}  
\usepackage{helvet} 
\usepackage{courier}  
\usepackage[hyphens]{url}  
\usepackage{graphicx} 
\usepackage{graphicx}  
\usepackage{mathptmx} 
\usepackage{times} 
\usepackage{amsmath} 
\usepackage{amssymb}  
\usepackage[protrusion=true,expansion=true]{microtype} 
\newcommand{\argmax}{\operatornamewithlimits{arg\,max}}
\usepackage{subfigure}
\usepackage{adjustbox}
\usepackage{url}
\usepackage[capitalize]{cleveref}
\usepackage{siunitx}
\usepackage{booktabs}
\usepackage{listings}
\usepackage{todonotes}
\usepackage[all]{nowidow}

\usepackage{pgfplots}
\pgfplotsset{compat=newest}
\pgfplotsset{every axis legend/.append style={%
cells={anchor=west}}
}
\usepgfplotslibrary{polar}
\usetikzlibrary{arrows}
\tikzset{>=stealth'}
\usepackage{pgf}
\usetikzlibrary{automata}
\usepgfplotslibrary{groupplots}

\usepackage[backend=bibtex,style=authoryear, maxbibnames=9,maxcitenames=2, natbib]{biblatex}
\AtEveryBibitem{
 	\clearfield{url}  
 	\clearfield{doi}  
\ifentrytype{inproceedings}{
 	\clearlist{address}
 	\clearlist{publisher}
 	\clearname{editor}
 	\clearlist{organization}
 	\clearfield{pages}  
 	\clearlist{location}
 	\clearfield{volume}
 }{}
 }

\AtBeginBibliography{\small}

\title{Point-Based Methods for Model Checking \\ in Partially Observable Markov Decision Processes }
\author{Maxime Bouton\\ {\tt boutonm@stanford.edu} \\ Stanford University \\ Stanford, CA \\
\And Jana Tumova \\  {\tt tumova@kth.se} \\ {KTH} Royal Institute of Technology\\
Stockholm, Sweden \\
\And Mykel J. Kochenderfer \\ {\tt mykel@stanford.edu} \\ Stanford University \\ Stanford, CA
}

\begin{document}

\maketitle

\begin{abstract}
Autonomous systems are often required to operate in partially observable environments. 
They must reliably execute a specified objective even with incomplete information about the state of the environment. 
We propose a methodology to synthesize policies that satisfy a linear temporal logic formula in a partially observable Markov decision process (POMDP). 
By formulating a planning problem, we show how to use point-based value iteration methods to efficiently approximate the maximum probability of satisfying a desired logical formula and compute the associated belief state policy. 
We demonstrate that our method scales to large POMDP domains and provides strong bounds on the performance of the resulting policy. 
\end{abstract}

\section{Introduction}

Designing decision making strategies for robotic systems in uncertain environments can be challenging. In many applications, the agent is equipped with sensors that are not capable of detecting all the relevant features of the environments. Sensors may not be able to detect objects through walls or directly measure the intentions of humans. Algorithms must generate strategies that are both efficient and reliable even in situations where all the information about the environment is not accessible. In addition, the resulting policies must exhibit strong guarantees on their performance.

A principled way to take into account both stochastic dynamics and state uncertainty is to model the environment as a partially observable Markov decision process (POMDP). 
The objective is often specified using an instantaneous reward function. The agent seeks to find a strategy that maximizes the expected accumulated reward over time. Defining reward functions can be very challenging and can lead to a value alignment problem, where the agent does not behave as expected~\citep{hadfield-menell2017}. 
Although existing planning algorithms can generate approximately optimal policies, it may not be straightforward how to interpret the performance of the policy through expected accumulated rewards.

In this work, we focus on the problem of synthesizing policies that achieve a desired objective expressed by a logical formula in a POMDP. We consider linear temporal logic (LTL)~\citep{pnueli77} as the framework for specifying the objective. LTL formulas can mathematically express objectives formulated in structured English~\citep{finucane2010}. In addition, we are interested in computing the probability of satisfying the desired formula when following the resulting policy. This problem is known as \textit{quantitative} model checking~\citep{baier2008}. In general, the problem of computing a policy that has the best probability of satisfying a logical formula in a POMDP is undecidable~\citep{chatterjee2013}. However, it is possible to derive approximate solutions to the problem with confidence bounds~\citep{hauskrecht2000}.

We propose a methodology to approximately solve quantitative model checking problems in POMDPs. 
We show that the problem of finding a policy maximizing the satisfaction of the objective can be formulated as a reward maximization problem. 
This consideration allows us to benefit from efficient approximate POMDP solvers, such as {SARSOP}~\citep{kurniawati2008}, to solve the original model checking problem. In addition, the bounds provided by the solver constitute strong guarantees on the performance of the resulting policy. 
We apply our methodology to classical POMDP domains and demonstrate that it can scale to larger environments than previous methods. We empirically verify that the probability of success of the policy is consistent with the upper and lower bounds provided by the solver. Finally, we compare the performance of point-based methods against previous work~\citep{norman2017}.

\section{Related Work}

Model checking in finite state Markov decision processes (MDPs) has been studied extensively and relies on two main solving strategies: value iteration and linear programs~\citep{baier2008,lahijanian2011}. These algorithms scale polynomially in the size of the MDP and efficient tools for probabilistic model checking can synthesize policies satisfying an LTL formula in MDPs with several millions states~\citep{kwiatkowska2011, dehnert2017}. However, these tools have little support for environments where the state is not observable, and current methods cannot scale to large POMDPs useful for robotics applications. 

The general problem of finding a policy satisfying an LTL formula in an infinite horizon POMDP is undecidable~\citep{chatterjee2013, chatterjee2015}. However, one can often compute approximate solutions by relaxing some aspects of the problem. A possible approach consists of restricting the space of policies to finite state controllers. This assumption can significantly reduce the search space. \citet{chatterjee2015} propose an exact algorithm relying on some heuristics to find policies satisfying a formula with probability 1. This algorithm has been used to synthesize policies in a drone surveillance problem~\citep{svorenova2015}. Other algorithms solve the quantitative model checking problem using parameter synthesis~\citep{junges2018} or a variant of value iteration~\citep{sharan2014}. The restriction to classes of policies with a limited number of internal states allows those approaches to scale to domains with thousands of states. However, in many applications, finite state policies might not be expressive enough to solve the problem. Instead, the policy must be represented as a mapping from a belief state (a distribution over states) to an action. 

\citeauthor{norman2017} addresses the problem of belief state planning with LTL specifications by discretizing the belief space and formulating an MDP over this space~\citep{norman2017}. In problems where the state space has more than a few dimensions, discretizing the belief space becomes intractable. We demonstrate that our method scales to problems with an order of magnitude more hidden states. Similarly, abstraction refinement methods were proposed to discretize the belief space in linear Gaussian POMDPs~\citep{haesaert2018}. Another approach for control in the belief space with LTL specifications linear Gaussian systems uses sampling based methods~\citep{vasile2016}. \citeauthor{wang2018} proposed an online search method to only explore belief points reachable from the current belief but their approach is limited to safe reachability objectives where the agent maximizes the probability of reaching a goal state while avoiding dangerous states~\citep{wang2018}. Alternative methods can check that a given belief state policy satisfies a safety or optimality criterion using barrier certificates but do not allow for policy synthesis~\citep{ahmadi2018}.

In this work, we propose a method to synthesize policies mapping belief states to actions with an LTL specification in a POMDP. We show that we can benefit from the advances in POMDP planning algorithms to solve model checking problems efficiently and avoid a naive discretization of the belief space. In contrast with previous work, we do not assume that the labels constituting the LTL formula are observable. In addition, our method handles stochastic observation models.

\section{Background}

This section reviews partially observable Markov decision processes and linear temporal logic.

\subsection{Partially Observable Markov Decision Processes}

Sequential decision making problems with state uncertainty can be modeled as partially observable Markov decision processes (POMDPs). They are formally defined by the tuple $(\mathcal{S}, \mathcal{A}, \mathcal{O},T,O,R,\gamma)$ where $\mathcal{S}$ is a finite state space, $\mathcal{A}$ a finite action space, $\mathcal{O}$ a finite observation space, $T$ a transition model, $O$ an observation model, $R$ a reward function, and $\gamma$ a discount factor. The transition model describes the probability of transitioning to a state $s'$ when taking an action $a\in\mathcal{A}$ in a state $s$: $T(s'\mid s, a) = \Pr(s'\mid s, a)$. When executing an action $a$ in a state $s$, the agent receives a scalar reward given by the function $R(s, a)$. The observation model represents the probability of observing $o\in\mathcal{O}$ while having executed action $a$ and being in state $s'$: $O(o \mid s', a) = \Pr(o \mid s', a)$.

During the decision process, the agent cannot sense the true state of the environment. Instead it maintains a belief that reflects its internal knowledge of the state. The \emph{belief state} is a probability distribution over all possible states, $b : \mathcal{S} \rightarrow [0,1]$, and $b(s)$ represents the probability of being in state $s$. In POMDPs with finite states, actions, and observations, the belief $b$ is updated after taking action $a$ and observing $o$ using the following equation:
\begin{equation}
b'(s') \propto O(o\mid s',a) \sum_s T(s'\mid a, s)b(s) 
\label{eq:update}
\end{equation}

A policy is a mapping from beliefs to actions. Given a policy $\pi$, an induced trajectory is a trajectory generated by an agent following $\pi$ from a given belief point. The solution to a POMDP is a policy $\pi^*$ that, if followed, maximizes the expected discounted sum of immediate rewards. The optimal policy can be extracted from the optimal belief action utility function $U^*(b, a)$ as follows:
\begin{equation}
    \pi^*(b) = \argmax_a U^*(b, a)
\end{equation}
where $U^*(b, a)$ represent the accumulated discounted reward obtained when following the optimal policy after taking action $a$ in belief $b$. We note $U^*(b) = \max_a U^*(b, a)$ the belief state utility function (also called value function).

When performing model checking, a convenient approach is to label the states of the POMDP and express the property we wish to verify in terms of these labels. 
The labels are atomic propositions that evaluate to true or false at a given state. We augment the definition of a POMDP with a finite set of atomic propositions $\Pi$, and $L$ a mapping, $L:\mathcal{S} \to 2^{\Pi}$, giving the set of atomic propositions satisfied at a given state. We do not assume that the labels are observable. The agent should infer the labels from the observations.

In this work, we focus on POMDPs with finite states, actions, and observations. We discuss possible extensions to continuous spaces in the conclusion.

\subsection{Linear Temporal Logic}

Linear Temporal Logic (LTL) is an extension to propositional logic with temporal operators. An LTL formula is built of atomic propositions according to the following grammar:
\begin{equation}
  \phi ::= p \mid \phi_1 \land \phi_2 \mid \phi_1 \lor \phi_2 \mid \lnot\phi \mid \mathsf{G} \phi \mid \mathsf{F} \phi \mid \phi_1 \mathsf{U} \phi_2 \mid \mathsf{X} \phi
  \label{eq:ltl}
\end{equation}
where $p$ is an atomic proposition, $\phi$, $\phi_1$, and $\phi_2$ are LTL formulas, $\lnot$ (negation), $\land$ (conjunction), and $\lor$ (disjunction) are logical operators, and $\mathsf{G}$ (globally), $\mathsf{F}$ (eventually), $\mathsf{U}$ (until), and $\mathsf{X}$ (next) are temporal operators~\citep{baier2008}. In this work we use LTL as a language to specify the objective of the problem. For example, safe-reachability objectives: ``avoid state $A$ and reach state $B$'' are specified by the formula $\lnot A \mathsf{U} B$, persistent tasks: ``keep visiting $A$'' are represented by the formula $\mathsf{G} \mathsf{F} A$. 

The satisfaction of an LTL formula is evaluated on an infinitely long trajectory in the environment. A labelling function maps each state of the environment to the set of atomic propositions holding in that state. The satisfaction of the formula can be verified by analyzing the sequence of atomic propositions generated by a trajectory. Even if the trajectory is continuous in time, the sequence of atomic propositions needs to be discrete.

\section{Proposed Approach}

This section presents our approach to solve the quantitative model checking problem using a POMDP formulation. We first demonstrate how to formulate a planning problem from a given model checking problem. Then, we explain how to approximately compute a policy that maximizes the probability of satisfying a given LTL formula. Finally we discuss how the convergence error of the solver can be used as a confidence interval on the resulting performance.

\subsection{Problem Formulation}

The problem of interest consists of computing the maximum probability of satisfying a given linear temporal logic formula $\phi$ when starting in an initial belief point $b$ in a POMDP.

Given a policy $\pi$, $\Pr^{\pi}(b \models \phi)$ represents the probability that a trajectory induced by $\pi$ starting from belief $b$ will satisfy the LTL formula $\phi$. The quantity we wish to compute is expressed as follows:

\begin{equation}
    {\textstyle\Pr}^{\max} (b \models \phi) = \max_{\pi} {\textstyle\Pr}^{\pi}(b \models \phi)
\end{equation}

Such problem is referred to as \textit{quantitative} model checking as opposed to \textit{qualitative} model checking, which consists of finding a policy satisfying the formula with probability~1~\citep{chatterjee2015}. In this work, the atomic propositions forming the LTL formula are defined over the states of the POMDP. Hence, the value of the atomic propositions is not observed by the agent. Instead, we will show that our formulation captures this information in the belief state.

\subsection{Reachability Problems}

Point-based value iteration methods can scale to POMDPs with many thousands states~\citep{kurniawati2008,shani2013}. Those solvers have been designed to solve reward maximization problems. We explain how to formulate reachability problems as reward maximization problems so we can use these solvers.

A reachability problem consists of computing the maximum probability of reaching a given set of states. If $B$ is a propositional formula then the reachability problem corresponds to computing $\Pr^{\max}(b \models \mathsf{F} B)$. For simplicity of the notation, we will also denote $B$, the set of states where the propositional formula expressed by $B$ holds true. A reachability problem can be interpreted as a planning problem where the goal is to reach the set $B$. This problem is addressed by defining the following reward function:
\begin{equation}
R_{\text{Reachability}}(s, a) = \begin{cases} 
    1 & \text{if } s \in B \\
    0 & \text{otherwise}
\end{cases}
\end{equation}
In addition, the states in the set $B$ are made terminal states and the initial value of $\Pr^{\max}(b \mid \mathsf{F} B)$ is initialized to 0 for any belief states. We can interpret the reachability problem as a reward maximization problem as follows:
\begin{equation}
{\textstyle\Pr}^{\max}(b \models \mathsf{F} B) = \max_{\pi} \mathbb{E}[\sum_t^\infty R_{\text{Reachability}}(s_t, \pi(b_t)) \mid s_0 \sim b]
\label{eq:reachability}
\end{equation}


The right side of this equation corresponds to solving a POMDP planning problem with a value-based method~\citep{kochenderfer2015}. The maximization is over the policy space. Note that in a POMDP, policies map belief states to actions rather than states to actions. The search problem becomes much harder than in MDPs and the value iteration algorithm can no longer scale. It has been proven that computing the maximum expected reward in a POMDP is undecidable~\citep{madani1999}. Instead, we will rely on approximate methods that scales to POMDP domains with tens of thousands of states. This step is discussed in depth in the section on approximate solution techniques. The next section discusses the generalization to any LTL formula.

\subsection{From LTL Satisfaction to Reachability}

\subsubsection{Product POMDPs} In this step, we define a new POMDP such that solving the original quantitative model checking problem reduces to a reachability problem in this model.

It is known that any LTL formula can be represented by a deterministic Rabin automaton~\citep{baier2008}, which can be defined as follows:

\textbf{Deterministic Rabin Automata (DRA): }
A deterministic Rabin automaton is a tuple $\mathcal{R} = (Q, \Pi, \delta, q_0, F)$ where $Q$ is a set of states, $\Pi$ a set of atomic propositions, $\delta : Q\times2^\Pi \rightarrow Q$ is a transition function, $q_0$ is an initial state, and $F$ is an acceptance condition: $F = \{ (L_1,K_1), \ldots, (L_k, K_k) \}$ where $L_i$ and $K_i$ are sets of states for all $i$. 

A trajectory of a Rabin automaton is an infinite sequence of states $\tau = q_0q_1\ldots$, where $q_{i+1} = \delta(q_i, \sigma)$ for an input $\sigma \in 2^\Pi$. We say that a trajectory is accepting if there exists $i$ such that: $\text{inf}(\tau) \cap K_i \neq \emptyset $ and $\text{inf}(\tau) \cap L_i = \emptyset $ where $\text{inf}(\tau)$ is the set of states visited infinitely often in the trajectory.
By converting the LTL formula into a DRA, we have a direct equivalence between accepting trajectories and trajectories satisfying the formula. 

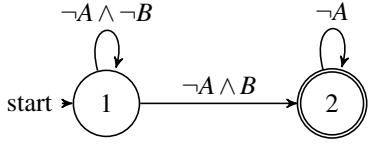
\begin{figure}
    \centering
            
              
                    
        
  \begin{tikzpicture}[->,>=stealth',auto, node distance = 3cm, semithick, scale=0.3]
    \node[initial, state] (A) {1};
    \node [state, accepting] (B) [right of=A] {2};
    
    \path (A) edge [loop above] node {$\lnot A \land \lnot B$} (A)
              edge node {$\lnot A \land B$} (B)
          (B) edge [loop above] node {$\lnot A$} (B);
  \end{tikzpicture}
  \caption{Illustration of an automaton generated by converting the LTL formula $\mathsf{G}\lnot A \land \mathsf{F} B$. State 2 must be visited infinitely often to satisfy the formula. Each propositional formula on the edges represents possibly multiple transitions labeled with the subsets of atomic propositions that satisfy the formula on the edge.}
  \label{fig:automata}
\end{figure}

In general, converting an LTL formula into a DRA results in a finite state machine with a number of states double exponential in the number of atomic propositions in the formula. In practice, a lot of heuristics can be used to reduce the number of states in the automaton to a reasonable number. We give an example of the automaton resulting from converting $\mathsf{G}\lnot A \land \mathsf{F} B$ in \cref{fig:automata}.

\textbf{Product POMDP: }
For a POMDP $\mathcal{P}$, and DRA $\mathcal{R}$, we define a product $\mathcal{P} \otimes	 \mathcal{R}$ as a POMDP: $\mathcal{P'} = (\mathcal{S}\times Q, \mathcal{A}, \mathcal{O}, T', O, L)$
where the state space is the Cartesian product of the state space of $\mathcal{P}$ and $\mathcal{R}$ and the transition function satisifies:
\begin{equation}
    T' ((s,q), a, (s', q')) = \begin{cases} 
        T(s, a, s') & \text{ if } q' = \delta(q, L(s)) \\
        0 & \text{ otherwise}
    \end{cases}
\end{equation}

all the other elements of the product are the same as in the original POMDP. In the product, some transitions are prevented by the automaton. We can notice that the transition function defined is no longer a probability distribution. In practice, we can add an additional sink state such that if $\delta(q, L(s)) = \emptyset$, the system transitions in the sink state with probability 1. The new transition function ensures that trajectories that end up in the sink state are not accepted by the automaton (they are violating the specification).

Let aside the model checking problem, the construction of the product POMDP can be interpreted as a principled way to augment the state space in order to account for temporal objective. In addition, one can note that this state space extension is not always necessary. For formulas involving only a single until ($\mathsf{U}$) or eventually ($\mathsf{F}$) temporal operators, the problem can be directly expressed as a reachability problem and does not require a state space augmentation.

\subsubsection{Maximal End Components}

The next step consists of identifying a set of states $B$ in the product POMDP, such that reaching a state in this set guarantees the satisfaction of the formula. We call those states \textit{success states}. 

From the definition of the DRA, we find that an infinitely long trajectory satisfying the formula must visit certain states infinitely often and others only finitely often. We first start to compute the sets of states that are visited infinitely often in the product POMDP, that is the maximal end component of a POMDP. More precisely, we need to find the maximal end components of the underlying MDP defined by ${(\mathcal{S}\times\mathcal{Q}, \mathcal{A}, T')}$. Starting from any state, with any policy, the agent will end up in a maximal end component if we consider infinitely long trajectories. Maximal end components can be computed by a graph algorithm that scales polynomially with the size of the state space~\citep{baier2008}. Once the end components have been found, we must identify the success states.

\textbf{Success States:}~\citep{baier2008} Given a product POMDP $\mathcal{P'}$, its underlying MDP is noted $\mathcal{M'}$. A state contained in a maximal end component $EC$ of $\mathcal{M'}$ is a success state if there exists an $i$ such that $K_i \in  EC$ and $L_i \notin EC$, where $K_i$ and $L_i$ results from the accepting conditions of the DRA used to form the product POMDP.

From the previous definition, we can conclude that from a success state, there is a probability of 1 of satisfying the LTL formula associated with the Rabin automaton. We can define a reachability reward function associated to the set of success states and compute the probability of success at a given belief point using \cref{eq:reachability}.

The first steps of the model checking approach (product POMDP and reduction to reachability) are identical for POMDPs and MDPs. They are independent of the structure of the observation space and are agnostic to partial observability. State uncertainty will play a role in the last step, which consists of solving the reachability problem.

\textbf{Theorem: }
Given a POMDP and an LTL formula $\phi$, the optimal value function of the product POMDP with the reachability reward function associated with the set of success states satisfies:
$U^*(b) = \Pr^{\max}(b \models \phi)$, where $b$ is a belief state in the product POMDP. In addition, there is a one to one mapping between the policy maximizing the value function in the product POMDP and the policy maximizing $\Pr(b \models \phi)$. 

\textbf{Proof Sketch:} The construction of the product POMDP, and the definition of success states give the following:
\begin{equation}
    {\textstyle\Pr}^{\max}_{\mathcal{P}}(b \models \phi) = {\textstyle\Pr}^{\max}_{\mathcal{P'}}(b \models \mathsf{F}B)
    \label{eq:product}
\end{equation}
where on both sides, $b$ is a belief of the product states, that is a belief over both the state of $\mathcal{P}$ and the state of the DRA associated with $\phi$, and $B$ is the set of success states in $\mathcal{P}'$. When updating the belief using \cref{eq:update}, the transition model from the product POMDP is used.
Finally, \cref{eq:reachability} holds from the construction of the reachability reward function and the definition of the belief state value function of a POMDP. 
More precisely, \cref{eq:reachability} can be proven by formulating a belief state MDP~\citep{kochenderfer2015} and use the equivalent result for MDPs~\citep{baier2008}.

The agent cannot observe whether it has reached an end component or not, but the belief state characterizes the confidence on whether or not it is in an end component.
Previous works often assume that the end components are observed, our algorithm allows to relax this assumption by maintaining a belief on both the state of the environment and the state of the automaton.

\subsection{Approximate Solution Techniques}\label{sec:solving}

The previous sections illustrated how to convert the quantitative model checking problem into a reward maximization problem. This section describes how to solve this problem using existing POMDP planning algorithms and how to interpret the convergence bounds with respect to the problem of interest. As we have shown, $\Pr^{\max}(b \models \phi)$ can be interpreted as a belief value function for a specific POMDP. This section  discusses how to compute such value function.

Solving POMDPs exactly is generally intractable~\citep{kochenderfer2015, madani1999}, however approximation techniques have been developed. Approximation methods rely on restricting the policy space, either by considering finite-state controllers or alpha vector representations. Previous work addressed the problem of finding finite state controllers~\citep{junges2018, chatterjee2015}. This paper focuses on alpha vector representations of the policy and the value function. An advantage of alpha vectors is that they can be used to represent both the policy and the value function. Hence, we can approximate the quantitative model checking problem and not only the policy synthesis problem. 

Alpha vectors are $|\mathcal{S}|$-dimensional vectors defining a linear function over the belief space. Given a set of alpha vectors $\Gamma = \{\alpha_1, \ldots, \alpha_n \}$, the value function is defined as follows: $U(b) = \max_{\alpha \in \Gamma} \alpha^\top b$

Point based Value Iteration (PBVI) algorithms are a family of POMDP solvers that involves applying a Bellman backup to a set of alpha vectors in order to approximate the optimal value function. \citet{shani2013} survey various PBVI methods. In this work, we used SARSOP~\citep{kurniawati2008}, which has shown state-of-the-art performance in terms of scalability. PBVI algorithms sample the belief space and compute an alpha vector associated to each belief point to approximate the value function at that point. SARSOP differs from other PBVI algorithms by relying on a tree search to explore the belief space. It maintains an upper and lower bound on the value function, which are used to guide the search close to optimal trajectories. The algorithm is given an initial belief point and only explores relevant regions of the belief space. That is, regions that can be reached from the initial belief point under optimality conditions.  

PBVI algorithms, often offer convergence guarantees specified in upper and lower bound on the value function. A precision parameter $\epsilon$ is provided and control the tightness of the convergence (by controlling the depth of the tree in SARSOP for example) which yields to:
\begin{equation}
    |\overline{U^*(b_0)} - \underline{U^*(b_0)}| < \epsilon
\end{equation}
Given a formula $\phi$, we have show how to build a product POMDP in which we have the equivalence between the value function $U^*(b)$ and ${\textstyle\Pr}^{\max}(b \models \phi)$. 
As a consequence, for a given precision parameter, we can directly translate the bounds on the value function in the product POMDP in terms of probability of success for our problem of quantitative model checking:
\begin{equation}
    |\overline{{\textstyle\Pr}^{\max}(b_0 \models \phi)} - \underline{{\textstyle\Pr}^{\max}(b_0 \models \phi)}| < \epsilon
\end{equation}
where $\overline{\Pr^{\max}(b_0 \models \phi)}$ is an upper bound over the actual probability of satisfaction, $\underline{\Pr^{\max}(b_0 \models \phi)}$ is a lower bound, and $b_0$ is the initial belief. With an infinite computation time, an arbitrary $\epsilon$ can be reached. However in practice only a minimum $\epsilon$ can be achieved within the computation budget. 
The original implementation of SARSOP relies on a discount factor. 
In this work, the discount factor is set to one such that the obtained value function matches exactly with the probability of satisfaction of the LTL formula.

The proposed methodology to solve quantitative model checking problems in POMDPs is agnostic to the planning algorithm. Although we focused the discussion on PBVI solvers, any belief state planner could be used. The strength of the guarantees are directly dependent on the choice of the underlying planning algorithm. For example, one could use the QMDP or FIB approximations to only compute an upper bound on the probability of success~\citep{hauskrecht2000}. Our implementation allows the user to easily choose the underlying algorithm among the one available in POMDPs.jl~\citep{egorov2017} a POMDP planning library.






\section{Experiments}

We evaluate our methodology on three discrete POMDP domains from the literature. The first one is a partially observable slippery grid world, the second one is the rock sample problem~\citep{smith2004}, and the third is a drone surveillance problem~\citep{svorenova2015}. Those domains have a grid world like structure and can easily be scaled to different size of state and observation spaces to evaluate the scalability of our approach. More details on the domains can be found in the available source code and in the supplementary material.


\subsubsection{Partially Observable Grid World}
This domain is an $n\times n$ grid with three labels: A, B, and C associated to some cells in the grid. The agent can choose to move left, right, up, and down. It reaches the desired cell with a probability of \num{0.7} and moves to another neighboring cell with equal probability otherwise. 
The agent receives a noisy observation of its position generated from a uniform distribution over the neighboring cells (vanish for distances greater than 1). The agent is initialized to a cell in the grid world with uniform probability. We investigated the following specifications:
\begin{itemize}
    \item $ \phi_1 = \lnot C \mathsf{U} A \land \lnot C \mathsf{U} B $: The agent must visit states A and B in any order while avoiding state C. This formula is a constrained reachability objective and does not require to form a product POMDP.
    \item $\phi_2 = \mathsf{G} \lnot C$: The agent must never visit state C.
\end{itemize}
The precision of the solver is set to \num{1e-2}.

\subsubsection{Drone Surveillance}
The drone surveillance problem is inspired by \citeauthor{svorenova2015}~\citep{svorenova2015}. An aerial vehicle must survey regions in the corners of a grid like environment while avoiding a ground agent. 
The drone can observe the location of the ground agent only if it is in its field of view delimited by a $3\times3$ area centered at the drone location.
We labeled the states as $A$ when the drone is in the bottom left corner, $B$ when it is in the top right corner, and \verb|det| when it can be detected by the ground agent (when it is on top of it). 
We analyzed one formula: $\lnot \verb|det| \mathsf{U} B$. The drone should eventually reach region B without being detected. 
Note that this is already a reachability objective and does not require the construction of a product POMDP. The precision is set to \num{1e-2}.

\subsubsection{Rock Sample}
The rock sample problem models a rover exploring a planet and tasked to collect interesting rocks. The environment consists of a grid world with rocks at a known location as well as an exit area. The rocks can be either good or bad and their status is not observable. The robot can move deterministically in each direction or choose to sample a rock (when on top of it), or use its long range sensor to check the quality of a rock. The long range sensor returns the true status of a rock with a probability decaying exponentially with the distance to the rock. 
The problem ends when the robot reaches the exit area, this state is labeled as $\verb|exit|$. In addition we defined two labels for situations when the robot pick a good rock or a bad rock respectively labeled $\verb|good|$ and $\verb|bad|$. This paper considers three different formulas:
\begin{itemize}
    \item $\phi_1 = \mathsf{G} \lnot \verb|bad|$ : This formula expresses that the robot should never pick up a bad rock. There exist a trivial policy that satisfies this formula which is to never pick up any rocks.
    \item $\phi_2 = \mathsf{F} \verb|good| \land \mathsf{F} \verb|exit|$: This formula expresses that the robot should eventually pick a good rock and eventually reach the exit. Since the exit is a terminal state, the robot must pick up a good rock before reaching the exit. This policy cannot be satisfied with a probability 1 since there is a possibility that all the rocks present are bad.
    \item $\phi_3 = \mathsf{F} \verb|good| \land \mathsf{F} \verb|exit| \land \mathsf{G} \lnot \verb|bad|$: This formula is a combination of the two previous specifications. In addition of bringing a good rock and reaching the exit the robot must not pick a bad rock. A video demonstrating the resulting strategy is provided in the supplementary material.
\end{itemize}
For this domain, the precision of the solver is set to \num{1e-3}.

\section{Results}

We applied the proposed methodology on different sizes of the proposed domains with different formulas. We use SARSOP as the underlying POMDP planning algorithm to solve the quantitative model checking problems. Note that our approach is agnostic to the choice of the planning algorithm and other methods could have been used. However, SARSOP is a good candidate for the task since it is one of the most scalable offline POMDP planners~\citep{kurniawati2008}. In addition, it provides strong bounds on the results, which can be translated into guarantees on the probability of success. 

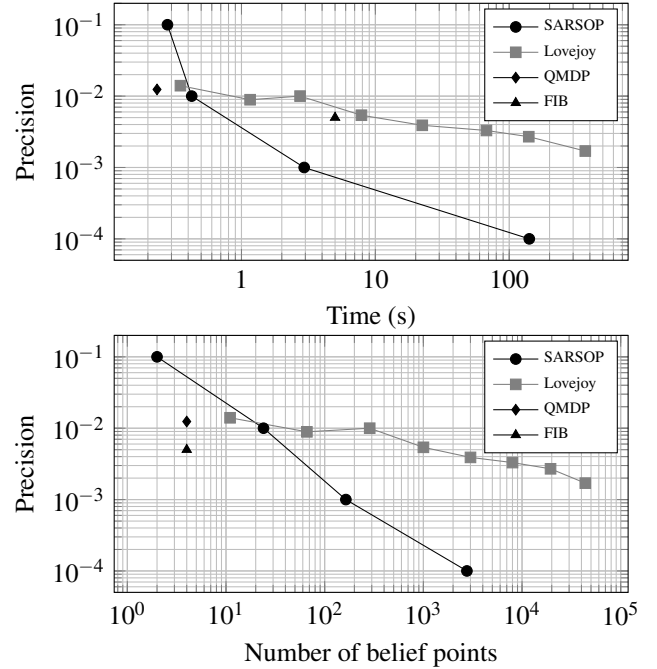
\begin{figure}[h!]
    \centering
    \input{precision_plot.tex}
    \caption{Illustration of the time precision trade-off for different algorithms providing upper bounds on the value function in a POMDP. Lovejoy is the algorithm used by \citeauthor{norman2017}. To compute the precision, we used the lower bound computed using SARSOP as a reference. The experiments are carried on a $3\times3$ partially observable grid world domain.}
    \label{fig:precision}
\end{figure}

We compared the performance of SARSOP with the algorithm used by \citet{norman2017}. It consists of computing an upper bound by discretizing the belief space and performing Bellman backups on each of the belief points~\citep{lovejoy1991}. The main drawback of this algorithm is that the belief space is high dimensional (12545 dimensions for the largest rock sample), and the size of the grid grows exponentially. \cref{fig:precision} illustrates the benefits of using SARSOP instead of the Lovejoy algorithm. The discretization scheme is controlled by a granularity parameter $m$, the bigger $m$ is, the more belief points are used. The Lovejoy line is obtained by varying $m$ from 1 to 8, while the SARSOP line is obtained by specifying different precision targets. In the log scale figure, we can see that it takes much longer time to reach a given precision using the Lovejoy algorithm than SARSOP. In addition, we can see the exponential growth of the number of belief points. As a reference we added the precision given by QMDP~\citep{littman1995} and FIB~\citep{hauskrecht2000} which are two algorithms to compute upper bound on the value of a POMDP. Point-based methods provide both an upper and desired bound and allow the user to specify the precision. Hence there is no need to use an abstraction refinement mechanism to choose the right granularity of the belief space as done in previous work~\citep{norman2017}.

\Cref{tab:results} summarizes the performance of our approach in solving different tasks. In each case, we report the lower bound on $\Pr^{\max}(b_0 \models \phi)$ as well as the precision $\epsilon$ described in previous sections. The upper bound is the sum of the two. In addition, we report the solving time, it takes into account both the time to compute the maximal end components in the product POMDP as well as the time taken by SARSOP to solve the problem. The MEC column reports the time needed to identify the success states and construct the product POMDP (if needed). To control the number of iterations used by SARSOP, we used a threshold on the precision, $\epsilon$ \textit{i.e.} after each iteration we check if the precision is lower than the threshold and return the policy and the probability of success if it is. The $|\Gamma|$ columns reports the number of belief points used by the point-based method.

\begin{figure}[h!]
    \centering
    \input{succ.tex}
    \caption{Estimate of the probability of success of a policy generated by SARSOP. We simulated \num{10000} episodes estimated the probability of success. We compare this result with the upper and lower bound provided by {SARSOP}.}
    \label{fig:mc_sim}
\end{figure}
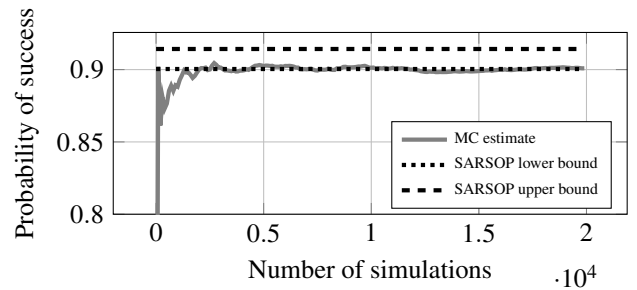

\begin{table}[t]
	\centering
	\caption{Performance of POMDP model checker.}
\begin{adjustbox}{max width=\columnwidth}
		\begin{tabular}{@{}lllSlll@{}}
		\toprule[1pt]
		Domain & $|\mathcal{S}|$ / $|\mathcal{A}|$ / $|\mathcal{O}|$ &  \text{LB} & {\text{$\epsilon$}} & \text{$|\Gamma|$} & \text{MEC (\SI{}{\second})} & \text{Time (\SI{}{\second})} \\
		\midrule
		\multicolumn{7}{l}{\textbf{PO Grid World}} \\
        $[10,10]$ $\phi_1$ & 101 / 4 / 101 & 0.904 & 9.9e-3 & 3452 & 0.64 & 207.2 \\
        $[10,10]$ $\phi_2$ &  & 0.0099 & 0 & 1 & 0.13 & 0.4 \\
        \midrule
	    \multicolumn{7}{l}{\textbf{Drone Surveillance}} \\
	    $[5,5]$ & 626 / 5 / 10 & 0.96 & 9e-3 & 4812 & 0.73 & 95.5 \\
	    $[5,5]$ (U)  & & 0.94 & 8e-3 & 4277 & 0.73 & 78.3 \\
	    $[7,7]$ (U)  & 2402 / 5 / 10  &  0.94 & 1.9e-2 & 41799 & 4.8 & 12587.5 \\
	    \midrule
		\multicolumn{7}{l}{\textbf{Rock Sample}} \\
	    $[4,4]$ $\phi_1$  & 65 / 7 / 3 &1.0 & 0.0 & 1 & 0.03 & 0.02 \\
	    $[4,4]$ $\phi_2$ & &  0.749 & 9.2e-5 & 13 & 0.09 & 0.3 \\
	    $[4,4]$ $\phi_3$ & &  0.744 & 2e-4 & 23 & 0.10 & 0.4 \\
	    $[5,5]$ $\phi_1$ & 201 / 8 / 3 &  1.0 & 0.0 & 1 & 0.19 &  0.11 \\
	    $[5,5]$ $\phi_2$ & & 0.879 & 2.8e-4 & 24 & 0.70 & 0.5 \\
	    $[5,5]$ $\phi_3$ & & 0.865 & 9e-4 & 56 & 0.70 & 0.8\\
	    $[7,7]$ $\phi_1$ & 12545 / 13 / 3 &  1.0  & 0.0 & 1 & 11.3 & 13.4 \\
	    $[7,7]$ $\phi_2$ & & 0.990 & 9e-4 & 378 & 50.6 & 77.5 \\
	    $[7,7]$ $\phi_3$ & &  0.979 & 9e-4 & 301 & 53.5  & 87.2\\
		\bottomrule[1pt]
	\end{tabular}%
\end{adjustbox}
	\label{tab:results}
\end{table}

We empirically verify the correctness of the bound provided by SARSOP by simulating the resulting policy in the partially observable grid world with the formula ${ \lnot C \mathsf{U} A \land \lnot C \mathsf{U} B }$. \cref{fig:mc_sim} illustrates the convergence of the estimated probability of success with the number of simulation of the policy. The probability of success is estimated using a Monte Carlo estimator. We can see that the estimated value converges towards the lower bound provided by SARSOP (dotted line). In this particular example, the value of the probability of success is around \num{0.90}. The gap between the upper and lower bound provided by the solver can be controlled with the precision, in expense of a longer time to solve. \cref{fig:mc_sim} shows that the resulting policy has an empirical performance consistent with the lower bound given by SARSOP.

\section{Discussion}

We have illustrated in the previous section that our approach scales to POMDP domains with many thousands states and supports different LTL specifications. We can see  from \cref{tab:results}, that the model checker is able to provide an approximate solution in a reasonable time. In contrast with previous work~\citep{svorenova2015, chatterjee2015}, solving a quantitative model checking problem instead of a qualitative problem allows us to find a policy even in cases where satisfiability cannot be guaranteed with probability 1. Moreover, our technique scales to larger state spaces.

In a few cases, the solver returned a policy with perfect precision in a very short time. This is the case for $\mathsf{G} \lnot C$ in grid world, and $\mathsf{G} \lnot \verb|bad|$ in rock sample. In those two cases, the probability of success can be directly extracted from the maximum end components. In the grid world example, the whole grid world is a maximal end component. The state space is fully connected under any policy because of the probabilistic transitions. As a consequence, there exists no trajectory that would not eventually visit the state C in an infinite time. This problem does not have any success states. In the rock sample problem, the transition is deterministic, there exist many trivial policies to not pick a bad rock. The robot can just stay idle, or reach the exit. In those two examples, computing the maximum end component and performing one iteration of SARSOP is enough to solve the model checking problem. 

For the large version of the drone surveillance problem, the computation reached a maximum memory limit on the size of the policy and was not able to reach the desired precision. Although this problem is smaller than rock sample, the belief space has a much denser support. The drone maintains a belief over the location of the agent outside its field of view. This characteristic of the belief space makes this problem harder to approximate~\citep{hsu2007}.

The solution provided by our approach is approximate. Although it provides mathematical bounds on the performance, it is not possible to compute the solution exactly. Reaching an arbitrary precision would require exploring the full belief space and take an infinite time. As a consequence, for smaller domains, approaches like the one proposed by \citeauthor{chatterjee2015} might be more suitable~\citep{chatterjee2015}. However, our approach does allow us to find approximate solutions in domains that were intractable for previous belief state approaches to model checking in POMDPs.
The formulation of the reward function in the product POMDP makes it a goal-oriented POMDP~\cite{kolobov2012}. Our methodology would allow one to replace the POMDP planner by a goal-oriented POMDP solver. 
It would require extending the algorithm from \citet{kolobov2012} to POMDPs. 
A comparison with traditional POMDP planners would be an interesting future direction. The dead end framework could be a useful theoretical framework to analyze the convergence of the solvers in the product POMDPs.

Contrary to previous work~\citep{norman2017}, we do not assume that the labels are observable. The computed policy maps a belief in the product space (POMDP state and automaton state) to an action. In problems where the automaton state is observable, our approach could still be applied and leverage this mixed observability assumption. This property would certainly help improve the results on the large drone surveillance problem. It has been shown that PBVI algorithms can scale to even larger domains when part of the state is fully observable~\citep{ong2009}.

\section{Conclusion}\label{sec:conclusion}

This paper proposed a methodology to solve quantitative model checking problems in POMDPs. Given an LTL formula and a POMDP model, our approach approximates the maximum probability of satisfying the formula as well as the corresponding belief state policy. We first convert the LTL formula into an automaton and construct a product POMDP between the automaton and the original POMDP model. By formulating a reward maximization problem, we have shown how to benefit from approximate POMDP planning algorithms to compute a solution to the model checking problem. Our method provides strong convergence bounds on the result. We have shown empirically that our approach applies to a variety of discrete POMDP domains, for different LTL formulas, and  scales to larger problem than previous belief state techniques~\citep{norman2017, svorenova2015}.
We provide a Julia package for POMDP model checking available at {\url{https://github.com/sisl/POMDPModelChecking.jl}}. 

The main limitation of the methodology is that it only applies to POMDPs with discrete state spaces. The two bottlenecks are the computation of the maximal end components and the choice of the planning algorithms. For some LTL formula, like constrained reachability~\citep{baier2008}, or if one is interested in policy synthesis only, the reward maximization problem can be formulated without having to compute maximal end components~\citep{sadigh2014}.  Our approach provides a flexible way to integrate LTL objectives in POMDP planning and allows to use any planning algorithm to allow a trade-off between convergence guarantees and scalability. Online POMDP planning algorithms could be used instead of PBVI methods to generate policies from an LTL objective at the price of lacking convergence guarantees. 


\section*{Acknowledgment}
This work was supported by the Honda Research Institute. The authors thank Sebastian Junges, Nils Jansen, and Emma Brunskill for their advice on the early stages of this work.

\printbibliography

\end{document}

%% file: precision_plot.tex
\begin{tikzpicture}[]
\begin{groupplot}[
group style={horizontal sep = 1.5cm, group size=1 by 2},
grid=both,
legend style={font=\tiny},
width = \columnwidth,
height = 5cm,
]
\nextgroupplot [ylabel = {Precision}, ymode = {log}, xlabel = {Time (\SI{}{\second})}, xmode = {log},
xticklabels = {0.2, 1, 10, 100}
]\addplot+[
black,
mark=*,
mark options={fill=black}
] coordinates {
(0.028, 0.1)
(0.0425, 0.01)
(0.294, 0.001)
(14.113, 0.0001)
};
\addlegendentry{SARSOP}
\addplot+[
gray,
mark=square*,
mark options={fill=gray}
]
coordinates {
(0.035, 0.014)
(0.116, 0.0089)
(0.274, 0.01)
(0.787, 0.0054)
(2.24, 0.0039)
(6.754, 0.0033)
(13.99, 0.0027)
(36.98, 0.0017)
};
\addlegendentry{Lovejoy}
\addplot+[
draw=none, 
black,
mark=diamond*,
mark options={fill=black}
] coordinates {
(0.0234, 0.0124)
};
\addlegendentry{QMDP}
\addplot+[
draw=none, 
black,
mark=triangle*,
mark options={fill=black}
] coordinates {
(0.5, 0.005)
};
\addlegendentry{FIB}
\nextgroupplot [ylabel = {Precision}, ymode = {log}, xlabel = {Number of belief points}, xmode = {log}]
\addplot+[
black,
mark=*,
mark options={fill=black}
]
coordinates {
(2.0, 0.1)
(24.0, 0.01)
(164.0, 0.001)
(2763.0, 0.0001)
};
\addlegendentry{SARSOP}
\addplot+[
gray,
mark=square*,
mark options={fill=gray}
]
coordinates {
(11.0, 0.014)
(66.0, 0.0089)
(286.0, 0.01)
(1001.0, 0.0054)
(3003.0, 0.0039)
(8008.0, 0.0033)
(19448.0, 0.0027)
(43758.0, 0.0017)
};
\addlegendentry{Lovejoy}
\addplot+[
draw=none, 
black,
mark=diamond*,
mark options={fill=black}
] coordinates {
(4.0, 0.0124)
};
\addlegendentry{QMDP}
\addplot+[
draw=none, 
black,
mark=triangle*,
mark options={fill=black}
] coordinates {
(4.0, 0.005)
};
\addlegendentry{FIB}
\end{groupplot}

\end{tikzpicture}

%% file: succ.tex
\begin{tikzpicture}[]
\begin{axis}[
height = 4cm,
width = \columnwidth,
ymin=0.8, 
legend pos = {south east}, ylabel = {Probability of success}, xlabel = {Number of simulations}, grid=both,
legend style={font=\tiny}
]\addplot+ [mark = {none}, gray, ultra thick]coordinates {
(1.0, 0.5)
(101.0, 0.9019607843137255)
(201.0, 0.8613861386138617)
(301.0, 0.8807947019867562)
(401.0, 0.8731343283582089)
(501.0, 0.8764940239043825)
(601.0, 0.8853820598006646)
(701.0, 0.8888888888888897)
(801.0, 0.8852867830423936)
(901.0, 0.889135254988914)
(1001.0, 0.8882235528942106)
(1101.0, 0.8929219600725946)
(1201.0, 0.8960066555740438)
(1301.0, 0.8993855606758836)
(1401.0, 0.8980028530670471)
(1501.0, 0.8981358189081234)
(1601.0, 0.8963795255930082)
(1701.0, 0.8942420681551115)
(1801.0, 0.8956714761376239)
(1901.0, 0.8980021030494224)
(2001.0, 0.9000999000998996)
(2101.0, 0.9010466222645103)
(2201.0, 0.9009990917347869)
(2301.0, 0.9013900955690711)
(2401.0, 0.8996669442131553)
(2501.0, 0.9000799360511594)
(2601.0, 0.9027671022290544)
(2701.0, 0.9045151739452257)
(2801.0, 0.9032833690221276)
(2901.0, 0.901447277739489)
(3001.0, 0.9013990672884739)
(3101.0, 0.9003868471953572)
(3201.0, 0.8994378513429112)
(3301.0, 0.8994548758328289)
(3401.0, 0.8991769547325098)
(3501.0, 0.8989149057681324)
(3601.0, 0.8995002776235422)
(3701.0, 0.8987034035656398)
(3801.0, 0.8987375065754873)
(3901.0, 0.8992824192721688)
(4001.0, 0.8993003498250881)
(4101.0, 0.8995611896635786)
(4201.0, 0.9002855782960493)
(4301.0, 0.8998140399814032)
(4401.0, 0.9007269422989546)
(4501.0, 0.9018214127054638)
(4601.0, 0.9026510212950883)
(4701.0, 0.9028073160357295)
(4801.0, 0.9031653477717614)
(4901.0, 0.9028967768257856)
(5001.0, 0.9024390243902437)
(5101.0, 0.9027832222657775)
(5201.0, 0.9027297193387159)
(5301.0, 0.9024896265560166)
(5401.0, 0.9026286560533141)
(5501.0, 0.902035623409669)
(5601.0, 0.901999285969297)
(5701.0, 0.9019642230796219)
(5801.0, 0.9022750775594617)
(5901.0, 0.9018976618095561)
(6001.0, 0.9025324891702783)
(6101.0, 0.9024909865617828)
(6201.0, 0.9026120606256056)
(6301.0, 0.9024119327197714)
(6401.0, 0.9020618556701028)
(6501.0, 0.9014149492463869)
(6601.0, 0.9013935171160254)
(6701.0, 0.9003282602208296)
(6801.0, 0.9001764187003823)
(6901.0, 0.9003187481889304)
(7001.0, 0.9001713796058265)
(7101.0, 0.9004505773021682)
(7201.0, 0.9005831713412928)
(7301.0, 0.9000273897562301)
(7401.0, 0.8994866252364219)
(7501.0, 0.8993601706211685)
(7601.0, 0.8998947645356489)
(7701.0, 0.9000259672812261)
(7801.0, 0.8998974621891834)
(7901.0, 0.9000253100480885)
(8001.0, 0.9006498375406158)
(8101.0, 0.9001481115773876)
(8201.0, 0.8997805413313822)
(8301.0, 0.8997831847747526)
(8401.0, 0.9008569388240897)
(8501.0, 0.9008468595624562)
(8601.0, 0.9010695187165775)
(8701.0, 0.9018616410020684)
(8801.0, 0.9019541013406047)
(8901.0, 0.9015951471579424)
(9001.0, 0.9015774272383911)
(9101.0, 0.9015600966820483)
(9201.0, 0.9016518148228649)
(9301.0, 0.9018490647172647)
(9401.0, 0.9021484790470105)
(9501.0, 0.902125868238266)
(9601.0, 0.9024161632993134)
(9701.0, 0.902597402597403)
(9801.0, 0.9020608039175678)
(9901.0, 0.9013330640274696)
(10001.0, 0.9013197360527884)
(10101.0, 0.9014056622450994)
(10201.0, 0.9008037639678498)
(10301.0, 0.9010871675402838)
(10401.0, 0.9009805806575658)
(10501.0, 0.9007808036564452)
(10601.0, 0.9011507262780604)
(10701.0, 0.9010465333582502)
(10801.0, 0.9006665432327342)
(10901.0, 0.9008438818565401)
(11001.0, 0.9011997818578447)
(11101.0, 0.9008286795172045)
(11201.0, 0.9010890912337093)
(11301.0, 0.9007255353034861)
(11401.0, 0.9009822838098579)
(11501.0, 0.9007998608937573)
(11601.0, 0.9005343906223064)
(11701.0, 0.9005298239617162)
(11801.0, 0.900016946280292)
(11901.0, 0.9001008233910267)
(12001.0, 0.8997667055490766)
(12101.0, 0.8996033713435789)
(12201.0, 0.8994427143091288)
(12301.0, 0.8988782311819205)
(12401.0, 0.8986453797774553)
(12501.0, 0.898416253399456)
(12601.0, 0.8983494683383594)
(12701.0, 0.8986773736419456)
(12801.0, 0.8985314794563355)
(12901.0, 0.898542861571849)
(13001.0, 0.8982464236271335)
(13101.0, 0.8982598076629527)
(13201.0, 0.8983487350401449)
(13301.0, 0.8982859720342814)
(13401.0, 0.8982987613788983)
(13501.0, 0.8983113612798109)
(13601.0, 0.8984708131157185)
(13701.0, 0.898554955480952)
(13801.0, 0.8987827851036069)
(13901.0, 0.8988634728816005)
(14001.0, 0.8987287530352807)
(14101.0, 0.8985959438377528)
(14201.0, 0.8986762427827071)
(14301.0, 0.8987554188225413)
(14401.0, 0.8989723649493121)
(14501.0, 0.8988415390980552)
(14601.0, 0.8987125051362814)
(14701.0, 0.898721262413276)
(14801.0, 0.8987974598027291)
(14901.0, 0.8988726345456991)
(15001.0, 0.8990134648713498)
(15101.0, 0.8990862137465236)
(15201.0, 0.8991580055255898)
(15301.0, 0.8994249117762391)
(15401.0, 0.899428645630439)
(15501.0, 0.8995613469229785)
(15601.0, 0.8991795923599538)
(15701.0, 0.8995032479938873)
(15801.0, 0.899632957853436)
(15901.0, 0.8998239215193061)
(16001.0, 0.899825021872266)
(16101.0, 0.899764004471494)
(16201.0, 0.9001975064806823)
(16301.0, 0.8998282419335037)
(16401.0, 0.8997683209364714)
(16501.0, 0.8996485274512185)
(16601.0, 0.899771111914227)
(16701.0, 0.8999521015447257)
(16801.0, 0.9001309367932393)
(16901.0, 0.9003076558987102)
(17001.0, 0.9001293965415829)
(17101.0, 0.9000116945386503)
(17201.0, 0.900069759330311)
(17301.0, 0.9004161368627898)
(17401.0, 0.9004137455464877)
(17501.0, 0.9007541995200549)
(17601.0, 0.9010339734121114)
(17701.0, 0.9010846232064171)
(17801.0, 0.9007976631839122)
(17901.0, 0.9008490671433358)
(18001.0, 0.9010109987779144)
(18101.0, 0.9011158987957137)
(18201.0, 0.9009449511042744)
(18301.0, 0.9012129821877385)
(18401.0, 0.901206390609717)
(18501.0, 0.9011998702842934)
(18601.0, 0.9013009353832914)
(18701.0, 0.9014009196877343)
(18801.0, 0.9014466546112122)
(18901.0, 0.9015977145275634)
(19001.0, 0.9014840543100726)
(19101.0, 0.901319233588106)
(19201.0, 0.9012602853869388)
(19301.0, 0.9010465236763041)
(19401.0, 0.9010411297804345)
(19501.0, 0.9012408983693984)
(19601.0, 0.9012345679012351)
(19701.0, 0.9010252766216635)
(19801.0, 0.9008685991314007)
(19901.0, 0.9009144809566889)
};
\addlegendentry{MC estimate}
\addplot+ [mark = {none}, ultra thick, black, dotted]coordinates {
(1.0, 0.900412)
(101.0, 0.900412)
(201.0, 0.900412)
(301.0, 0.900412)
(401.0, 0.900412)
(501.0, 0.900412)
(601.0, 0.900412)
(701.0, 0.900412)
(801.0, 0.900412)
(901.0, 0.900412)
(1001.0, 0.900412)
(1101.0, 0.900412)
(1201.0, 0.900412)
(1301.0, 0.900412)
(1401.0, 0.900412)
(1501.0, 0.900412)
(1601.0, 0.900412)
(1701.0, 0.900412)
(1801.0, 0.900412)
(1901.0, 0.900412)
(2001.0, 0.900412)
(2101.0, 0.900412)
(2201.0, 0.900412)
(2301.0, 0.900412)
(2401.0, 0.900412)
(2501.0, 0.900412)
(2601.0, 0.900412)
(2701.0, 0.900412)
(2801.0, 0.900412)
(2901.0, 0.900412)
(3001.0, 0.900412)
(3101.0, 0.900412)
(3201.0, 0.900412)
(3301.0, 0.900412)
(3401.0, 0.900412)
(3501.0, 0.900412)
(3601.0, 0.900412)
(3701.0, 0.900412)
(3801.0, 0.900412)
(3901.0, 0.900412)
(4001.0, 0.900412)
(4101.0, 0.900412)
(4201.0, 0.900412)
(4301.0, 0.900412)
(4401.0, 0.900412)
(4501.0, 0.900412)
(4601.0, 0.900412)
(4701.0, 0.900412)
(4801.0, 0.900412)
(4901.0, 0.900412)
(5001.0, 0.900412)
(5101.0, 0.900412)
(5201.0, 0.900412)
(5301.0, 0.900412)
(5401.0, 0.900412)
(5501.0, 0.900412)
(5601.0, 0.900412)
(5701.0, 0.900412)
(5801.0, 0.900412)
(5901.0, 0.900412)
(6001.0, 0.900412)
(6101.0, 0.900412)
(6201.0, 0.900412)
(6301.0, 0.900412)
(6401.0, 0.900412)
(6501.0, 0.900412)
(6601.0, 0.900412)
(6701.0, 0.900412)
(6801.0, 0.900412)
(6901.0, 0.900412)
(7001.0, 0.900412)
(7101.0, 0.900412)
(7201.0, 0.900412)
(7301.0, 0.900412)
(7401.0, 0.900412)
(7501.0, 0.900412)
(7601.0, 0.900412)
(7701.0, 0.900412)
(7801.0, 0.900412)
(7901.0, 0.900412)
(8001.0, 0.900412)
(8101.0, 0.900412)
(8201.0, 0.900412)
(8301.0, 0.900412)
(8401.0, 0.900412)
(8501.0, 0.900412)
(8601.0, 0.900412)
(8701.0, 0.900412)
(8801.0, 0.900412)
(8901.0, 0.900412)
(9001.0, 0.900412)
(9101.0, 0.900412)
(9201.0, 0.900412)
(9301.0, 0.900412)
(9401.0, 0.900412)
(9501.0, 0.900412)
(9601.0, 0.900412)
(9701.0, 0.900412)
(9801.0, 0.900412)
(9901.0, 0.900412)
(10001.0, 0.900412)
(10101.0, 0.900412)
(10201.0, 0.900412)
(10301.0, 0.900412)
(10401.0, 0.900412)
(10501.0, 0.900412)
(10601.0, 0.900412)
(10701.0, 0.900412)
(10801.0, 0.900412)
(10901.0, 0.900412)
(11001.0, 0.900412)
(11101.0, 0.900412)
(11201.0, 0.900412)
(11301.0, 0.900412)
(11401.0, 0.900412)
(11501.0, 0.900412)
(11601.0, 0.900412)
(11701.0, 0.900412)
(11801.0, 0.900412)
(11901.0, 0.900412)
(12001.0, 0.900412)
(12101.0, 0.900412)
(12201.0, 0.900412)
(12301.0, 0.900412)
(12401.0, 0.900412)
(12501.0, 0.900412)
(12601.0, 0.900412)
(12701.0, 0.900412)
(12801.0, 0.900412)
(12901.0, 0.900412)
(13001.0, 0.900412)
(13101.0, 0.900412)
(13201.0, 0.900412)
(13301.0, 0.900412)
(13401.0, 0.900412)
(13501.0, 0.900412)
(13601.0, 0.900412)
(13701.0, 0.900412)
(13801.0, 0.900412)
(13901.0, 0.900412)
(14001.0, 0.900412)
(14101.0, 0.900412)
(14201.0, 0.900412)
(14301.0, 0.900412)
(14401.0, 0.900412)
(14501.0, 0.900412)
(14601.0, 0.900412)
(14701.0, 0.900412)
(14801.0, 0.900412)
(14901.0, 0.900412)
(15001.0, 0.900412)
(15101.0, 0.900412)
(15201.0, 0.900412)
(15301.0, 0.900412)
(15401.0, 0.900412)
(15501.0, 0.900412)
(15601.0, 0.900412)
(15701.0, 0.900412)
(15801.0, 0.900412)
(15901.0, 0.900412)
(16001.0, 0.900412)
(16101.0, 0.900412)
(16201.0, 0.900412)
(16301.0, 0.900412)
(16401.0, 0.900412)
(16501.0, 0.900412)
(16601.0, 0.900412)
(16701.0, 0.900412)
(16801.0, 0.900412)
(16901.0, 0.900412)
(17001.0, 0.900412)
(17101.0, 0.900412)
(17201.0, 0.900412)
(17301.0, 0.900412)
(17401.0, 0.900412)
(17501.0, 0.900412)
(17601.0, 0.900412)
(17701.0, 0.900412)
(17801.0, 0.900412)
(17901.0, 0.900412)
(18001.0, 0.900412)
(18101.0, 0.900412)
(18201.0, 0.900412)
(18301.0, 0.900412)
(18401.0, 0.900412)
(18501.0, 0.900412)
(18601.0, 0.900412)
(18701.0, 0.900412)
(18801.0, 0.900412)
(18901.0, 0.900412)
(19001.0, 0.900412)
(19101.0, 0.900412)
(19201.0, 0.900412)
(19301.0, 0.900412)
(19401.0, 0.900412)
(19501.0, 0.900412)
(19601.0, 0.900412)
(19701.0, 0.900412)
(19801.0, 0.900412)
(19901.0, 0.900412)
};
\addlegendentry{SARSOP lower bound}
\addplot+ [mark = {none}, ultra thick, black, dashed]coordinates {
(1.0, 0.914258)
(101.0, 0.914258)
(201.0, 0.914258)
(301.0, 0.914258)
(401.0, 0.914258)
(501.0, 0.914258)
(601.0, 0.914258)
(701.0, 0.914258)
(801.0, 0.914258)
(901.0, 0.914258)
(1001.0, 0.914258)
(1101.0, 0.914258)
(1201.0, 0.914258)
(1301.0, 0.914258)
(1401.0, 0.914258)
(1501.0, 0.914258)
(1601.0, 0.914258)
(1701.0, 0.914258)
(1801.0, 0.914258)
(1901.0, 0.914258)
(2001.0, 0.914258)
(2101.0, 0.914258)
(2201.0, 0.914258)
(2301.0, 0.914258)
(2401.0, 0.914258)
(2501.0, 0.914258)
(2601.0, 0.914258)
(2701.0, 0.914258)
(2801.0, 0.914258)
(2901.0, 0.914258)
(3001.0, 0.914258)
(3101.0, 0.914258)
(3201.0, 0.914258)
(3301.0, 0.914258)
(3401.0, 0.914258)
(3501.0, 0.914258)
(3601.0, 0.914258)
(3701.0, 0.914258)
(3801.0, 0.914258)
(3901.0, 0.914258)
(4001.0, 0.914258)
(4101.0, 0.914258)
(4201.0, 0.914258)
(4301.0, 0.914258)
(4401.0, 0.914258)
(4501.0, 0.914258)
(4601.0, 0.914258)
(4701.0, 0.914258)
(4801.0, 0.914258)
(4901.0, 0.914258)
(5001.0, 0.914258)
(5101.0, 0.914258)
(5201.0, 0.914258)
(5301.0, 0.914258)
(5401.0, 0.914258)
(5501.0, 0.914258)
(5601.0, 0.914258)
(5701.0, 0.914258)
(5801.0, 0.914258)
(5901.0, 0.914258)
(6001.0, 0.914258)
(6101.0, 0.914258)
(6201.0, 0.914258)
(6301.0, 0.914258)
(6401.0, 0.914258)
(6501.0, 0.914258)
(6601.0, 0.914258)
(6701.0, 0.914258)
(6801.0, 0.914258)
(6901.0, 0.914258)
(7001.0, 0.914258)
(7101.0, 0.914258)
(7201.0, 0.914258)
(7301.0, 0.914258)
(7401.0, 0.914258)
(7501.0, 0.914258)
(7601.0, 0.914258)
(7701.0, 0.914258)
(7801.0, 0.914258)
(7901.0, 0.914258)
(8001.0, 0.914258)
(8101.0, 0.914258)
(8201.0, 0.914258)
(8301.0, 0.914258)
(8401.0, 0.914258)
(8501.0, 0.914258)
(8601.0, 0.914258)
(8701.0, 0.914258)
(8801.0, 0.914258)
(8901.0, 0.914258)
(9001.0, 0.914258)
(9101.0, 0.914258)
(9201.0, 0.914258)
(9301.0, 0.914258)
(9401.0, 0.914258)
(9501.0, 0.914258)
(9601.0, 0.914258)
(9701.0, 0.914258)
(9801.0, 0.914258)
(9901.0, 0.914258)
(10001.0, 0.914258)
(10101.0, 0.914258)
(10201.0, 0.914258)
(10301.0, 0.914258)
(10401.0, 0.914258)
(10501.0, 0.914258)
(10601.0, 0.914258)
(10701.0, 0.914258)
(10801.0, 0.914258)
(10901.0, 0.914258)
(11001.0, 0.914258)
(11101.0, 0.914258)
(11201.0, 0.914258)
(11301.0, 0.914258)
(11401.0, 0.914258)
(11501.0, 0.914258)
(11601.0, 0.914258)
(11701.0, 0.914258)
(11801.0, 0.914258)
(11901.0, 0.914258)
(12001.0, 0.914258)
(12101.0, 0.914258)
(12201.0, 0.914258)
(12301.0, 0.914258)
(12401.0, 0.914258)
(12501.0, 0.914258)
(12601.0, 0.914258)
(12701.0, 0.914258)
(12801.0, 0.914258)
(12901.0, 0.914258)
(13001.0, 0.914258)
(13101.0, 0.914258)
(13201.0, 0.914258)
(13301.0, 0.914258)
(13401.0, 0.914258)
(13501.0, 0.914258)
(13601.0, 0.914258)
(13701.0, 0.914258)
(13801.0, 0.914258)
(13901.0, 0.914258)
(14001.0, 0.914258)
(14101.0, 0.914258)
(14201.0, 0.914258)
(14301.0, 0.914258)
(14401.0, 0.914258)
(14501.0, 0.914258)
(14601.0, 0.914258)
(14701.0, 0.914258)
(14801.0, 0.914258)
(14901.0, 0.914258)
(15001.0, 0.914258)
(15101.0, 0.914258)
(15201.0, 0.914258)
(15301.0, 0.914258)
(15401.0, 0.914258)
(15501.0, 0.914258)
(15601.0, 0.914258)
(15701.0, 0.914258)
(15801.0, 0.914258)
(15901.0, 0.914258)
(16001.0, 0.914258)
(16101.0, 0.914258)
(16201.0, 0.914258)
(16301.0, 0.914258)
(16401.0, 0.914258)
(16501.0, 0.914258)
(16601.0, 0.914258)
(16701.0, 0.914258)
(16801.0, 0.914258)
(16901.0, 0.914258)
(17001.0, 0.914258)
(17101.0, 0.914258)
(17201.0, 0.914258)
(17301.0, 0.914258)
(17401.0, 0.914258)
(17501.0, 0.914258)
(17601.0, 0.914258)
(17701.0, 0.914258)
(17801.0, 0.914258)
(17901.0, 0.914258)
(18001.0, 0.914258)
(18101.0, 0.914258)
(18201.0, 0.914258)
(18301.0, 0.914258)
(18401.0, 0.914258)
(18501.0, 0.914258)
(18601.0, 0.914258)
(18701.0, 0.914258)
(18801.0, 0.914258)
(18901.0, 0.914258)
(19001.0, 0.914258)
(19101.0, 0.914258)
(19201.0, 0.914258)
(19301.0, 0.914258)
(19401.0, 0.914258)
(19501.0, 0.914258)
(19601.0, 0.914258)
(19701.0, 0.914258)
(19801.0, 0.914258)
(19901.0, 0.914258)
};
\addlegendentry{SARSOP upper bound}
\end{axis}

\end{tikzpicture}